\documentclass[11pt,a4paper]{article}
\usepackage[hyperref]{acl2021}
\usepackage{times}
\usepackage{latexsym}

\usepackage{microtype}
\usepackage{amssymb}
\usepackage{amsmath}
\usepackage{graphicx}
\usepackage{paralist}
\usepackage{lipsum}
\aclfinalcopy 

\title{Towards User-Driven Neural Machine Translation}
 
\author{Huan Lin\textsuperscript{1,2}~~~
Liang Yao\textsuperscript{3}~~~
Baosong Yang\textsuperscript{3}~~~
Dayiheng Liu\textsuperscript{3}~~~
Haibo Zhang\textsuperscript{3}\\ 
\textbf{Weihua Luo\textsuperscript{3}~~~
Degen Huang\textsuperscript{4}~~~
Jinsong Su\textsuperscript{1,2,5\thanks{~~Jinsong Su is the corresponding author. This work was done when Huan Lin was interning at DAMO Academy, Alibaba Group.}}}\\
  \textsuperscript{1}School of Informatics, Xiamen University \\
  \textsuperscript{2}Institute of Artificial Intelligence, Xiamen University\\
  \textsuperscript{3}Alibaba Group~~~\textsuperscript{4}Dalian University of Technology~~~
  \textsuperscript{5}Pengcheng Lab, Shenzhen\\
  
  \texttt{\small{huanlin@stu.xmu.edu.cn}} \\
  \texttt{\small{\{yaoliang.yl,yangbaosong.ybs,liudayiheng.ldyh,zhanhui.zhb\}@alibaba-inc.com}}\\
  \texttt{\small{weihua.luowh@alibaba-inc.com}}~~~
  \texttt{\small{huangdg@dlut.edu.cn}}~~~
  \texttt{\small{jssu@xmu.edu.cn}}

  }

\date{}

\begin{document}
\maketitle

\begin{abstract}
A good translation should not only translate the original content semantically, but also incarnate personal traits of the original text. For a real-world neural machine translation (NMT) system, these user traits (e.g., topic preference, stylistic characteristics and expression habits) can be preserved in user behavior (e.g., historical inputs). However, current NMT systems marginally consider the user behavior due to: 1) the difficulty of modeling user portraits in zero-shot scenarios, and 2) the lack of user-behavior annotated parallel dataset. To fill this gap, we introduce a novel framework called user-driven NMT. Specifically, a cache-based module and a user-driven contrastive learning method are proposed to offer NMT the ability to capture potential user traits from their historical inputs under a zero-shot learning fashion. Furthermore, we contribute the first Chinese-English parallel corpus annotated with user behavior called \textbf{UDT-Corpus}. Experimental results confirm that the proposed user-driven NMT can generate user-specific translations.
\footnote{We release our source code and the associated benchmark at \url{https://github.com/DeepLearnXMU/User-Driven-NMT}.}

\end{abstract}

\section{Introduction} \label{sec:intro}

\begin{figure}[!t]
\centering
\includegraphics[width=0.85\linewidth]{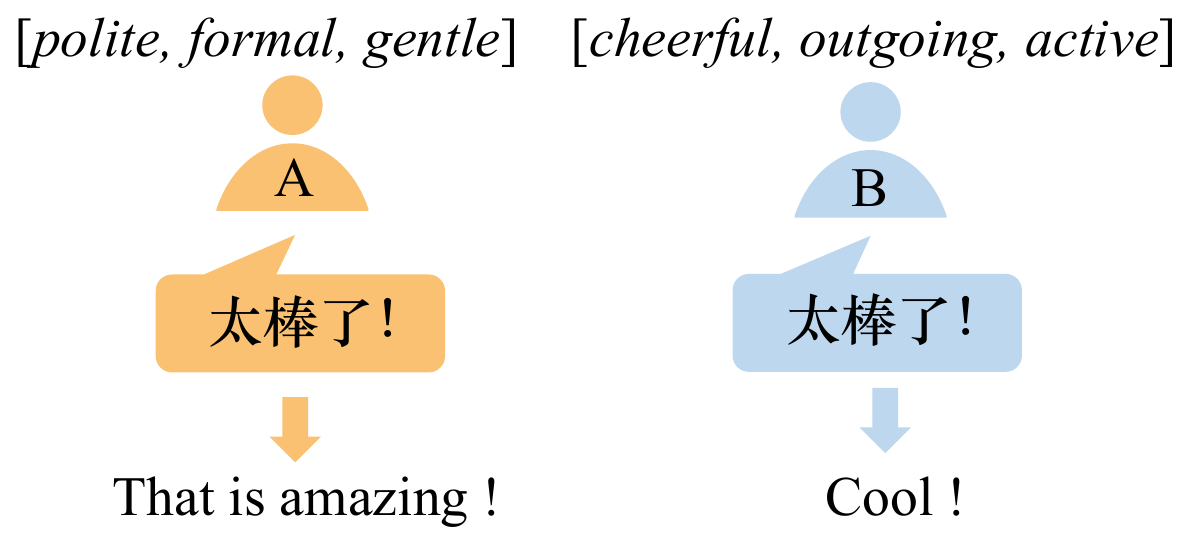}
\caption{
\label{fig_example}
An example in which user traits leads to synonymous yet stylistically different translations.}
\end{figure}

In recent years, neural machine translation (NMT) models~\citep{nips/SutskeverVL14,emnlp/LuongPM15,vaswani2017attention} have shown promising quality and thus increasingly attracted users. 
When drawing on a translation system, every user has his own traits, including topic preference, stylistic characteristics, and expression habits, which can be implicitly embodied in their behavior, e.g., the historical inputs of these users. 
A good translation should implicitly mirror user traits rather than merely translate the original content, as the example shown in Figure \ref{fig_example}. 
However, current NMT models are mainly designed for the semantic transformation between the source and target sentences regardless of subtle traits with respect to user behavior. It can be said that the effect of user behavior on translation modeling is still far from utilization, which, to some extent, limits the applicability of NMT models in real-world scenarios.

More recently, several studies have shown that the prominent signals in terms of personal characteristics can be served as inductive biases and reflected in translation results using domain adaptation approaches, such as personality~\citep{emnlp/MirkinNBP15}, gender~\citep{eacl/WintnerMSRP17}, and  politeness~\citep{naacl/SennrichHB16}. However, previously explored signals characterize users from a single dimension, which insufficiently represent fine-grained user traits. 
Furthermore, \citet{acl/MichelN18} pay their attention to personalized TED talk translation, in which they train a speaker-specific bias to revise the prediction distribution. 
In contrast with these studies, our work investigates a more realistic online scenario: a real-world MT system serves extensive users, where the user-behavior annotated data covering all users is unavailable.  
Previous methods~\citep{emnlp/MirkinNBP15, acl/MichelN18} require the users in the training set and the test set to be consistent, therefore can not deal with this zero-shot issue. 

Starting from this concern, we explore user-driven NMT that generates personalized translations for users unseen in the training dataset according to their behavior. Specifically, we choose the historical inputs to represent user behavior since they can not only be easily obtained in the real-world scenarios, but also reflect the topic preference, stylistic characteristic, and context of user. 
Moreover, compared with pre-defined or user-specific labels, historical inputs can be updated with current source sentences, which is also in line with realistic scenario. 

In this work, we propose a novel framework for this task, where the NMT model is equipped with a cache module to restore and update historical inputs. Besides, in order to further transfer the traits from the seen users to the unseen ones, we design a regularization framework based on contrastive learning~\citep{acl/BoseLC18,acl/YangCLS19}, which forces our model to decrease the divergence between translations of similar users while increasing the diversity on dissimilar users.

In order to further train and assess the proposed framework, we construct a new \textbf{U}ser-\textbf{D}riven Machine \textbf{T}ranslation dataset called \textbf{UDT-Corpus}. This corpus consists of 6,550 users with totally 57,639 Chinese sentences collected from a real-world online MT system.
Among them, 17,099 Chinese sentences are annotated with their English translations by linguistic experts according to the user-specific historical inputs. Experimental results demonstrate that the proposed framework facilitates the translation quality, and exactly generates diverse translations for different users. 

To summarize, major contributions of our work are four-fold:
\begin{compactitem}
\item 
We introduce and explore user-driven NMT task that leverages user behavior to enhance translation model. We hope our study can attract more attention to explore techniques on this topic.
\item
We propose a novel framework for user-driven NMT based on cache module and contrastive learning, which is able to model user traits in zero-shot scenarios. 
\item
We collect UDT-Corpus and make it publicly available, which may contribute to the subsequent researches in the communities of NMT and user-driven models. 
\item
Extensive analyses indicate the effectiveness of our work and verify that NMT can profit from user behavior to generate diverse translations conforming to user traits. 
\end{compactitem}

\section{Related Work}
This section mainly includes the related studies of personalized machine translation, cache-based NMT and contrastive learning for NMT.
\paragraph{Personalized Machine Translation}
Recently, some researchers have employed domain adaptation~\citep{naacl/ZhangSKMCD19,acl/GururanganMSLBD20,coling/YaoYZCL20} to generate personalized translations. For example, \citet{emnlp/MirkinNBP15} show that the translation generated by the SMT model has an adverse effect on the prediction of author personalities, demonstrating the necessity of personalized machine translation. Furthermore, \citet{naacl/SennrichHB16} control the politeness in the translation by adding a politeness label on the source side. \citet{eacl/WintnerMSRP17} explore a gender-personalized SMT system that retains the original gender traits. These domain labels represent users in 
single dimension separately, which are insufficient to distinguish large-scale users in a fine-grained way.
The most correlated work to ours is \citet{acl/MichelN18} which introduces a speaker-specific bias into the conventional NMT model. However, these methods are unable to deal with users unseen at the training time. Different from them, user-driven NMT can generate personalized translations for these unseen users in a zero-shot manner.

\paragraph{Cache-Based Machine Translation}
Inspired by the great success of cache on language modeling~\citep{pami/KuhnM90,csl/Goodman01,interspeech/FedericoBC08}, \citet{emnlp/NepveuLLF04} propose a cache-based adaptive SMT system. \citet{tiedemann2010} explore a cache-based translation model that fills the cache with bilingual phrase pairs extracted from previous sentence pairs in a document. 
\citet{bertoldi2013} use a cache mechanism to achieve online learning in phrase-based SMT. 
\newcite{emnlp/GongZZ11}, \newcite{coling/KuangXLZ18}, and 
\newcite{tacl/TuLSZ18} further exploit cache-based approaches to leverage contextual information for document-level machine translation. 
Contrast with the document-level NMT that learns to capture contextual information, our study aims at modeling user traits, such as, topic preference, stylistic characteristics, and expression habits. Moreover, historical inputs of user has relatively fewer dependencies than the contexts used in document-level translation. 

\paragraph{Contrastive Learning for NMT}\label{sec:contra}
Contrastive learning has been extensively applied in the communities of computer vision and natural language processing due to its effectiveness and generality on self-supervised learning~\citep{emnlp/VaswaniZFC13,nips/MnihK13,aaai/LiuS15,acl/BoseLC18}. Towards raising the ability of NMT in capturing global dependencies, \citet{emnlp/WisemanR16} first introduce contrastive learning into NMT, where the ground-truth translation and the model output are considered as the positive and contrastive samples, respectively. \citet{acl/YangCLS19} construct contrastive examples by deleting words from ground-truth translation to reduce word omission errors in NMT. 
Contrast to these studies, we employ contrastive learning to create broader learning signals for our user-driven NMT model, where the prediction  distribution of translations with respect to similar users and dissimilar users are considered as positive and contrastive samples, respectively. Thus, our model can better transfer the knowledge of the seen users to the unseen ones.

\section{User-Driven Translation Dataset}
In order to build a user-driven NMT system, we construct a new dataset called UDT-Corpus containing 57,639 inputs of 6,550 users, 17,099 among them are Chinese-to-English translation examples. 

\subsection{Data Collection and Preprocessing}
We collect raw examples from Alibaba Translate\footnote{\url{https://www.aliyun.com/product/ai/base_alimt}} which contain the user inputs and the  translations given by the translation system.

For data preprocessing, we first anonymize data and perform data deduplication within each user. Then, we utilize a pre-trained n-gram language model KenLM\footnote{\url{https://github.com/kpu/kenlm}.} to filter out translation examples with low-quality source data. Moreover, we remove such pairs whose source sentence is shorter than 2 words or longer than 100 words.

\subsection{Data Annotation}
In the corpus, we represent each translation example as a triplet $\langle X^{(u)},Y^{(u)},H^{(u)} \rangle$, where $H^{(u)}$ is the historical inputs of the user $u$, $X^{(u)}$ is the current source sentence and $Y^{(u)}$ is the target translation sentence annotated with $H^{(u)}$. To obtain such a triplet, we first sequentially sample up to 10 source sentences which are the historical inputs of each user. Then, for the given historical inputs, we collect their followed source input paired with the pseudo translation given by the translation system. Afterwards, we assign these historical inputs and the current input pairs to two professional annotators and ask them to revise the pseudo translation according to the source sentence and historical inputs. 
Specifically, we first ask one of them to annotate and the other to evaluate, and then resolve annotation disagreements by reviewing. 
During annotation, 91.8\% of the original data are revised. Moreover, annotators are asked to record whether their revision is affected by user history. The result shows that 76.25\% of the sentences are impacted. 

\section{User-Driven NMT Framework}
In this section,
we first give a brief description about the problem formulation of user-driven NMT,
and then introduce our proposed framework in detail. 
We choose Transformer~\cite{vaswani2017attention} as the basic NMT model due to its competitive performance. In fact, our framework is transparent and applicable to other NMT models.

\begin{figure}[!t]
\centering
\includegraphics[width=1\linewidth]{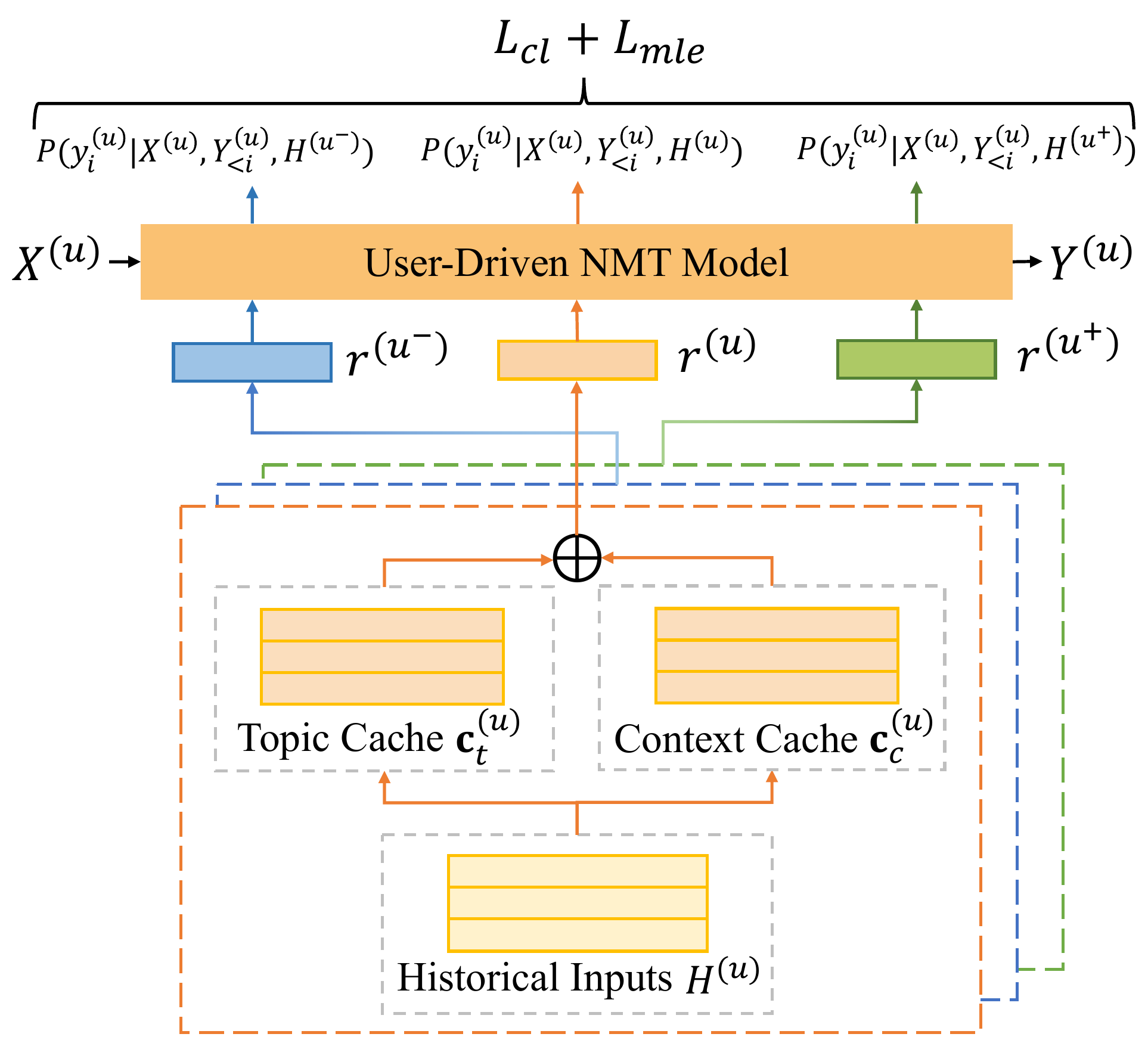}
\caption{
\label{fig_model}
The architecture of our user-driven NMT model. We use the topic cache and context cache to capture the long-term and short-term user traits for user $u$ from corresponding historical inputs $H^{(u)}$, respectively. Then, we combine the representations of two caches to get a user behavior representation $r^{(u)}$, which is fed into the NMT model for personalized translation. Furthermore, we use contrastive learning involving similar user $u^+$ and dissimilar user $u^-$ to increase the translation diversity among different users. 
}
\end{figure}

Figure \ref{fig_model} illustrates the basic framework of the proposed user-driven NMT.
Most typically, 
we equip the NMT model with two user-specific caches to exploit user behavior for better translation (See Section \S~\ref{sec:cache}). 
Besides, 
we augment the conventional NMT training objective with contrastive learning, which allows the model to learn translation diversity across users (See Section \S~\ref{sec:contra}).

\subsection{Problem Formulation}\label{sec:problem}
Given the source sentence $X$ and the previously generated words $Y_{<i}=y_1,...,y_{i-1}$, 
the conventional NMT model with parameter $\theta$ predicts the current target word $y_{i}$ by $P\left(y_{i}|X,Y_{<i};\theta \right)$. 
As a significant extension of conventional NMT, 
user-driven NMT with parameter $\theta$ aims to model $P\left(y_{i}^{(u)}|X^{(u)},Y_{<i}^{(u)},u;\theta \right)$, 
that is, 
generates the translation that can reflect the traits of user $u$. 
Unlike previous studies~\cite{emnlp/MirkinNBP15, acl/MichelN18} only caring for generating translations for users seen at the training time,
our user-driven NMT mainly focuses on a more realistic online MT scenario, 
where the users for testing are unseen in the training dataset. 
Moreover, the conventional domain adaptation methods can not be directly applied to this zero-shot scenario.

\subsection{Cache-based User Behavior Modeling}\label{sec:cache}
Due to the advantages of cache mechanism on dynamic representations~\citep{emnlp/GongZZ11,coling/KuangXLZ18,tacl/TuLSZ18}, 
we equip the conventional Transformer-based NMT model with two user-specific caches to leverage user behavior for NMT: 
1) \emph{topic cache} $\textbf{c}^{(u)}_t$ that aims at capturing the global and long-term traits of user $u$; and 
2) \emph{context cache} $\textbf{c}^{(u)}_c$, 
which is introduced to capture the short-term traits from the recent source inputs of user $u$. 
During this process, we focus on the following three operations on cache: 

\paragraph{Cache Representation} 
In order to facilitate the efficient computation of the user behavior encoded by our caches, 
we define each cache as an embedding sequence of keywords. 
We first calculate TF-IDF values of input words, 
and then extract words with TF-IDF weights higher than a predefined threshold to represent user behavior. 

Note that the calculation of TF-IDF value of a word mainly depends on its frequency in the \emph{document} and inverse document frequency in the \emph{corpus}. 
Since two caches play different roles in the user-driven NMT model, we identify keywords for two caches based on different definitions of ``document'' and ``corpus''. 
Specifically, when constructing topic cache $\textbf{c}^{(u)}_t$, 
we treat the historical inputs $H^{(u)}$ of the user $u$ as the ``document'' and the historical inputs $H^{(u)}$ of all users $U$ as the ``corpus'', 
then define topic cache $\textbf{c}^{(u)}_t$ as an embedding sequence of historical keywords. 
Unlike the topic cache, 
for context cache $\textbf{c}^{(u)}_c$, we individually consider the current source sentence $X^{(u)}$ and historical inputs $H^{(u)}$ as the TF-IDF ``document'' and ``corpus'', defining $\textbf{c}^{(u)}_c$ as an embedding sequence of current keywords. 

Besides, in the real-world MT scenario, 
there exists a large number of users without any historical input. 
For these users, we find the most similar user according to the cosine similarity based on their TF-IDF bag-of-word representations of topic keywords, and initialize the corresponding topic cache with that of the most similar user.

\paragraph{Updating Caches}
When using an online MT system, 
users often continuously input multiple sentences. 
Thus, 
our caches should be dynamically updated to ensure the accurate encoding of user behavior. 

To update topic cache, 
we first recalcualte the TF-IDF values of all historical input words,
so as to redetermine the keywords stored in this cache.
As for context cache, 
we consider it as a \emph{filter window} sliding across historical inputs, 
and apply \emph{first-in-first-out} rule to replace its earliest keywords with the recently input ones.

\paragraph{Reading from Caches}
During the translation of the NMT model, 
we perform a gating operation on $\textbf{c}^{(u)}_t$ and $\textbf{c}^{(u)}_c$, 
producing a vector $r^{(u)}$ that reflects user behavior as follows:
\begin{align}
&r^{(u)} = \mathbf{\alpha}\overline{c}_t^{(u)} + (1-\mathbf{\alpha})\overline{c}_c^{(u)} \\
&\mathbf{\alpha} = \text{Sigmoid}(\mathbf{W}_t\overline{c}_t^{(u)}+\mathbf{W}_r\overline{c}_c^{(u)}),\\
&\overline{c}_t^{(u)}=\textrm{MeanPooling}\left[\textbf{c}^{(u)}_t\right],\\
&\overline{c}_c^{(u)}=\textrm{MeanPooling}\left[\textbf{c}^{(u)}_c\right],
\end{align}
where both $W_t$ and $W_r$ are learnable parameter matrices. 
Then, we directly add $r^{(u)}$ into the embedding sequence of original current source sentence $X^{(u)}$, forming a source embedding sequence with user behavior as follows:
\begin{equation}
    \hat{X}^{(u)}=\{x^{(u)}_i+r^{(u)}\}_{1\leq i<|X^{(u)}|}.
\end{equation}
Finally, the NMT model is fed with $\hat{X}(u)$ to generate the translation for $u$.
Due to the limitation of pages, we omit the detailed descriptions of the NMT model. Please refer to \citet{vaswani2017attention} for the details.

\subsection{Model Training with a Contrastive Loss}\label{sec:contra}

Given training instances $\langle X^{(u)},Y^{(u)},H^{(u)}\rangle$, we train the user-driven NMT model using the following objective function:
\begin{equation}
    \mathcal{L}=\mathcal{L}_{\textrm{mle}}+\mathcal{L}_{\textrm{cl}}.
\end{equation}
Here, $\mathcal{L}_{\textrm{mle}}$ is the maximum likelihood translation loss extended from the conventional NMT training objective. Formally, it is defined as: 
\begin{equation}
    \mathcal{L}_{\textrm{mle}}=\sum_{i}-\log P(y_{i}^{(u)}|X^{(u)},Y_{<i}^{(u)},H^{(u)};\theta).
\end{equation}
$\mathcal{L}_{\textrm{cl}}$ is a triplet-margin-based constrastive loss, which allows the NMT model to learn the translation diversity across users. 
 

Specifically, for an input sentence, an ideal user-driven NMT model should be able to generate translations with non-divergent user traits for similar users, while producing translations with diverse user traits for dissimilar users. 
However, using only $\mathcal{L}_{\textrm{mle}}$ cannot guarantee this since it separately considers each training instance during the model training. 
To deal with this issue, for each training instance $\langle X^{(u)},Y^{(u)},H^{(u)}\rangle$, we first determine the most similar user $u^{+}$ according to the cosine similarity based on their bag-of-keyword representations, and randomly select a user without any same keyword as the dissimilar user $u^-$ of $u$. 
Finally, using historical inputs of $u^+$ and $u^-$, we construct several pseudo training instances to define $\mathcal{L}_{\textrm{cl}}$ as follows: 
\begin{align}
    \mathcal{L}_{\textrm{cl}}= &\sum_{u\in U}\textrm{max}[ d(X^{(u)},Y^{(u)},H^{(u)},H^{(u^+)})  \\ & - d(X^{(u)},Y^{(u)},H^{(u)},H^{(u^-)})+\eta,0 ], \notag
\end{align}
where $ d\left(X^{(u)},Y^{(u)},H^{(u)},H^{(u^+)}\right)$
\begin{align}\label{eqt_distance}
    &=||\frac{1}{|Y^{(u)}|}\sum_i\log P\left (y^{(u)}_{i}|X^{(u)},Y^{(u)}_{<i},H^{(u)}\right) \notag \\
    &-\frac{1}{|Y^{(u)}|}\sum_i\log P\left (y^{(u)}_{i}|X^{(u)},Y^{(u)}_{<i},H^{(u^+)}\right)||^2 
\end{align}
and $\eta$ is a predefined threshold, which is set to 2 in our experiments. Here, we omit the definition of $ d\left(X^{(u)},Y^{(u)},H^{(u)},H^{(u^-)}\right)$, which is similar to $ d\left(X^{(u)},Y^{(u)},H^{(u)},H^{(u^+)}\right)$.

Formally, $\mathcal{L}_{\textrm{cl}}$ will encourage the NMT model to minimize the prediction difference between the training instances $\langle X^{(u)},Y^{(u)},H^{(u)}\rangle$ and $\langle X^{(u)}, Y^{(u)},H^{(u^+)}\rangle$,
and maximize the difference between the training instances $\langle X^{(u)},Y^{(u)},H^{(u)}\rangle$ and $\langle X^{(u)}, Y^{(u)},H^{(u^-)}\rangle$.
In this way, the NMT model can not only exploit pesudo training instances, 
but also produce more consistent translations with user traits.

\section{Experiments}
In this section, we carry out several groups of experiments to investigate the effectiveness of our proposed framework on UDT-Corpus. 


\subsection{Setup}
We develop the user-driven NMT model based on Open-NMT Transformer~\citep{klein-etal-2017-opennmt}, 
and adopt a two-stage strategy to train this model:
we first pre-train a Transformer-based NMT model on the WMT2017 Chinese-to-English dataset,
and then fine-tune this model to our user-driven NMT model using UDT-Corpus. 

\paragraph{Datasets} 
The WMT2017 Chinese-to-English dataset is composed of the News Commentary v12, UN Parallel Corpus v1.0, and CWMT corpora, with totally 25M parallel sentences. To fine-tune our model, we split UDT-Corpus into training, validation and test set, respectively. 
Table \ref{tab_dataset} provides more detailed statistics of these datasets. 
To improve the efficiency of model training, we train the model using only parallel sentences with no more than 100 words. 
Following common practices, we employ byte pair encoding~\citep{sennrich2015neural} with 32K merge operations to deal with all sentences. 

\begin{table}
\centering
\scalebox{0.93}{
\begin{tabular}{l|r|r|r}
\hline 
&\textbf{Train} & \textbf{Dev} & \textbf{Test} \\ 
\hline
\#user                & 5,350  & 600   & 600 \\
\#historical input    & 33,441 & 3,629 & 3,470\\
\#current sentence pairs       & 14,006 & 1,557 & 1,536\\
\hline
\end{tabular}}
\caption{Dataset for fine-tuning experiments.}
\label{tab_dataset}
\end{table}

\paragraph{Training Details} 
Following \citet{vaswani2017attention}, we use the following hyper-parameters: the word embedding dimension is set to 512, the hidden layer dimension is 2048, the layer numbers of both encoder and decoder are set to 6, and the number of attention heads is set to 8. 
Besides, we use 4 GPUs for training. At the pre-training stage, we employ the Adam optimizer with $\beta_2$ = 0.998. We use the batch size of 16,384 tokens and pre-train the model for 200,000 steps. 
Particularly, we adopt the dropout strategy~\citep{srivastava2014dropout} with rate 0.1 to enhance the robustness of our model. 
When fine-tuning the model, we keep the other settings consistent with the pre-training stage, but reduce the batch size to 2048 tokens and fine-tune the model with early-stopping strategy.

\paragraph{Evaluation} 
We assess the translation quality with two metrics: one is case-insensitive BLEU \citep[\emph{mteval-v13a.pl},][]{papineni2002bleu}\footnote{\url{https://github.com/moses-smt/mosesdecoder/blob/master/scripts/generic/multi-bleu.perl}} and the other is METEOR\footnote{\url{https://github.com/cmu-mtlab/meteor}}~\citep{denkowski2011meteor}.

\subsection{Baselines}
We represent our user-driven NMT model as \textbf{UD-NMT} and compare it with the following baselines:
\begin{compactitem}
    \item \textbf{TF}. It is a Transformer-based NMT model pre-trained on the WMT2017 corpus. This model yields 24.61 BLEU score on WMT2017 Chinese-to-English translation task, which is comparable with reported results in~\citep{emnlp/WanYWZCZC20,acl/ZhouYWWC20}, which makes our subsequent experiments convincing.
    \item \textbf{TF-FT}. This model is also a Transformer-based NMT model that is further fine-tuned on the parallel sentences of UDT-Corpus.
    \item \textbf{TF-FT + PesuData}. This model is a variant of \emph{TF-FT}. When constructing it, we pair historical inputs with their translations produced by our online translation system, forming additional data for fine-tuning \emph{TF-FT}.
    \item \textbf{TF-FT + ConcHist}~\citep{discomt/TiedemannS17}. In this model, we introduce user behavior into \emph{TF-FT} by concatenating each input sentence with several historical inputs. We mark all tokens in historical inputs with a special prefix to indicate that they are additional information. 
    \item \textbf{TF-FT + UserBias}~\citep{acl/MichelN18}. It introduces user-specific biases to refine softmax-based predictions of Transformer NMT model. We change it to a zero-shot method similar to~\citep{wmt/FarajianTNF17} since~\citep{acl/MichelN18} can not be directly applied to our scenario. In particular, we replace the user ID in the test set with that of the most similar user in the training set.
\end{compactitem}
Note that the first two baselines, e.g., \emph{TF} and \emph{TF-FT}, 
are conventional NMT models without exploiting user behavior.

\subsection{Effect of Cache Sizes}
Since cache size directly determines the utility of user behavior, we investigate its effect on the performance of \textbf{UD-NMT}. 
We denote the sizes of topic cache and context cache as $s_t$ and $s_c$ for simplicity. 

\begin{figure}[!t]
\centering
\includegraphics[width=1\linewidth]{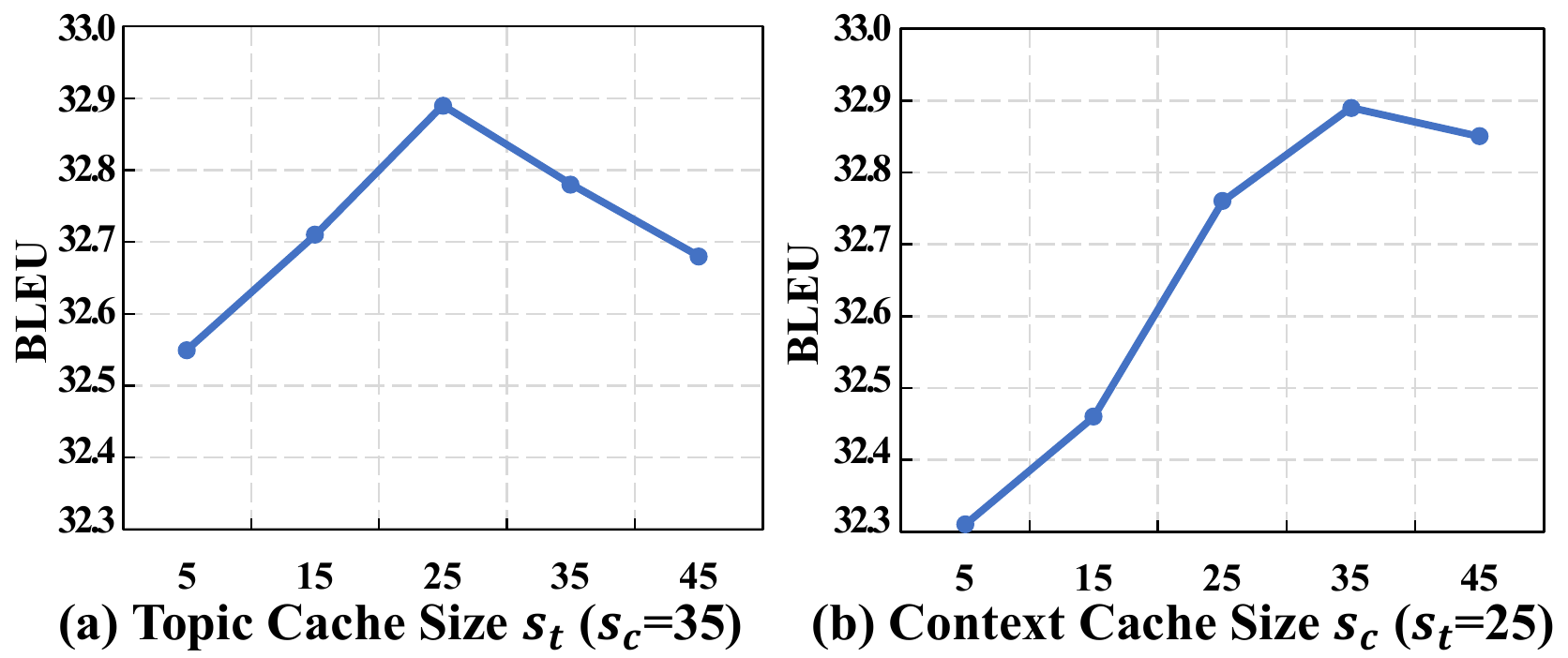}
\caption{
\label{fig_cache_size}
Effects of cache size on translation quality. 
}
\end{figure}

Figure \ref{fig_cache_size} lists the performance of our model with different $s_t$ and $s_c$ on validation set. We observe that $s_t$ larger than 25 and $s_c$ larger than 35 do not lead to significant improvements. 
For this result, we speculate that small cache sizes are unable to capture sufficient user behavior for NMT. 
However, since the number of keywords are limited, larger cache sizes only bring limited information gain. Therefore, we directly use $s_t$ = 25 and $s_c$ = 35 in the subsequent experiments.

\begin{table}
\centering
\begin{tabular}{l|c|c}
\hline \textbf{Model} & \textbf{BLEU} & \textbf{METEOR} \\ 
\hline
\multicolumn{3}{c}{\emph{w/o user behavior}} \\
\hline
TF                     & 27.52   & 44.05\\
TF-FT                  & 28.61   & 45.35\\
TF-FT + PesuData       & 29.02   & 45.40\\
\hline
\multicolumn{3}{c}{\emph{w/ user behavior}} \\
\hline
TF-FT + ConcHist       & 30.85   & 46.08 \\
TF-FT + UserBias       & 31.36   & 46.79 \\
UD-NMT                 & \textbf{32.35}   & \textbf{48.20}\\
\hline
\end{tabular}
\caption{Main results on UDT-Corpus. \emph{``w/o''}, \emph{``w/''} denote ``without'' and ``with'', respectively.
}
\label{tab_main_results}
\end{table}

\subsection{Main Results} \label{sec:main_exp}

\begin{table*} 
\centering
\scalebox{0.92}{
\begin{tabular}{l|c|c|c|c|c|c}
\hline \textbf{Model}            &\textbf{BLEU$\uparrow$} &\textbf{METEOR$\uparrow$} &\textbf{s-BLEU$\uparrow$} &\textbf{d-BLEU$\uparrow$} &\textbf{s-Sim.$\downarrow$} &\textbf{d-Sim.$\downarrow$}\\ \hline
UD-NMT                           & \textbf{32.35~}        & \textbf{48.20~}          & \textbf{32.17~}          & \textbf{32.23~}          & 93.18~                     & 80.10~ \\
w/o topic cache                  & 31.88$^{\dagger}$      & 48.00~                   & --                       & --                       & --                         & --  \\
w/o context cache                & 31.86$^{\dagger}$      & 47.84$^{\dagger}$        & 31.94$^{\dagger}$        & 31.58$^{\dagger}$        & \textbf{88.61~}     & \textbf{69.32~}  \\
w/o similar user initialization  & 32.02~                 & 48.14~                   & 31.86$^{\dagger}$        & 31.13$^{\ddagger}$       & 93.54$^{\dagger}$                     & 80.16~~\\
w/o contrastive learning         & 32.00~                 & 48.09~                   & 31.88$^{\dagger}$        & 31.94~                   & 93.49$^{\dagger}$          & 81.59$^{\dagger}$  \\
\hline
\end{tabular}}
\caption{Ablation Study. $\uparrow$: higher is better, $\downarrow$: lower is better. Since the user similarity is calculated based on the topic keywords, the model can not find similar user and dissimilar user without it. Thus \emph{w/o topic cache} does not have the s-BLEU, s-Sim., d-BLEU and d-Sim.. $\ddagger$/$\dagger$: indicates the drop of translation quality is statistically significant comparing to ``UD-NMT'' ($p$\textless 0.01/0.05).}
\label{tab_ablation}
\end{table*}

From Table \ref{tab_main_results}, 
we observe that our \emph{UD-NMT} model consistently outperforms all baselines in terms of two metrics. 
Moreover, we draw several interesting conclusions:

1) All NMT models leveraging user behavior surpass vanilla models, including \emph{TF}, \emph{TF-FT}, 
showing that user behavior is useful for NMT.

2) \emph{UD-NMT} exhibits better than \emph{TF-FT + PesuData},
which uses the same training data as ours. 
The underlying reason is that \emph{UD-NMT} can leverage user traits to generate better translations. 

3) Although both \emph{TF-FT + UserBias} and \emph{UD-NMT} exploit user behavior for NMT, \emph{UD-NMT} achieves better performance than \emph{TF-FT + UserBias} without introducing extra parameters. 
This result demonstrates the advantage of cache on modeling user behavior than introducing user-specific biases into model parameters.

\subsection{Ablation Study}

To explore the effectiveness of different components in our model,
we further compare \emph{UD-NMT} with its several variants, as shown in Table \ref{tab_ablation}.

Particularly,
we propose to evaluate translations using the following variant metrics: 
\textbf{s-BLEU}, \textbf{s-Sim.}, \textbf{d-BLEU} and \textbf{d-Sim.}.
When using s-BLEU, we replace the topic cache of current user with that of his most similar user. Keeping the same current input, we calculate the BLEU score with ground-truth as reference and the translation for this similar user as hypothesis. 
As for s-Sim., we adopt the same strategy as s-BLEU, but use the translation for original user as reference to evaluate the BLEU score. 
In other words, s-BLEU and d-BLEU assesses the translation quality given unsuitable user. Therefore, higher s-BLEU and d-BLEU indicates better model robustness, while  
s-BLEU and d-BLEU measures how much the translation changes given different user. Thus lower s-Sim. and d-Sim. show larger translation diversity.

Our conclusions are shown as follows: 

1) \emph{w/o topic cache}. To build this variant, we remove topic cache from our model. The result in Line 2 indicates that removing topic cache leads to a performance drop, suggesting that topic cache is useful for modeling user behavior.

2) \emph{w/o context cache}. Unlike the above variant, we only use topic cache to represent user traits in this variant. According to the results shown in Line 3, 
we observe that this change results in a significant performance decline of our model, 
demonstrating that context cache also effectively captures user behavior for NMT. 
However, the translation diversity among users increases since the model will not be affected by the context cache in this variant, which is the same between different users when calculating s-Sim. and d-Sim..

3) \emph{w/o similar user initialization}. In this variant, we do not initialize topic caches of the users without historical inputs using that of the most similar users. From Line 4, we observe that the performance of our model degrades without similar user initialization. 

4) \emph{w/o contrastive learning}. 
In this variant, we remove the contrastive learning from the whole training objective to inspect the performance change of our model. As shown in Line 4, the performance of our model drops, proving that the contrastive learning is important for the training of our model. 

Moreover, we can infer from \textbf{Column} 6 and 7 that our model can generate diverse translations. Specifically, the translations of dissimilar users has larger diversity than that of similar ones. 
Furthermore, we conclude that our model is robust, since it still performs well when we replace the topic cache of current user with those of other users (See \textbf{Column} 4 and 5).

\begin{figure*}[!t]
\centering
\includegraphics[width=1\linewidth]{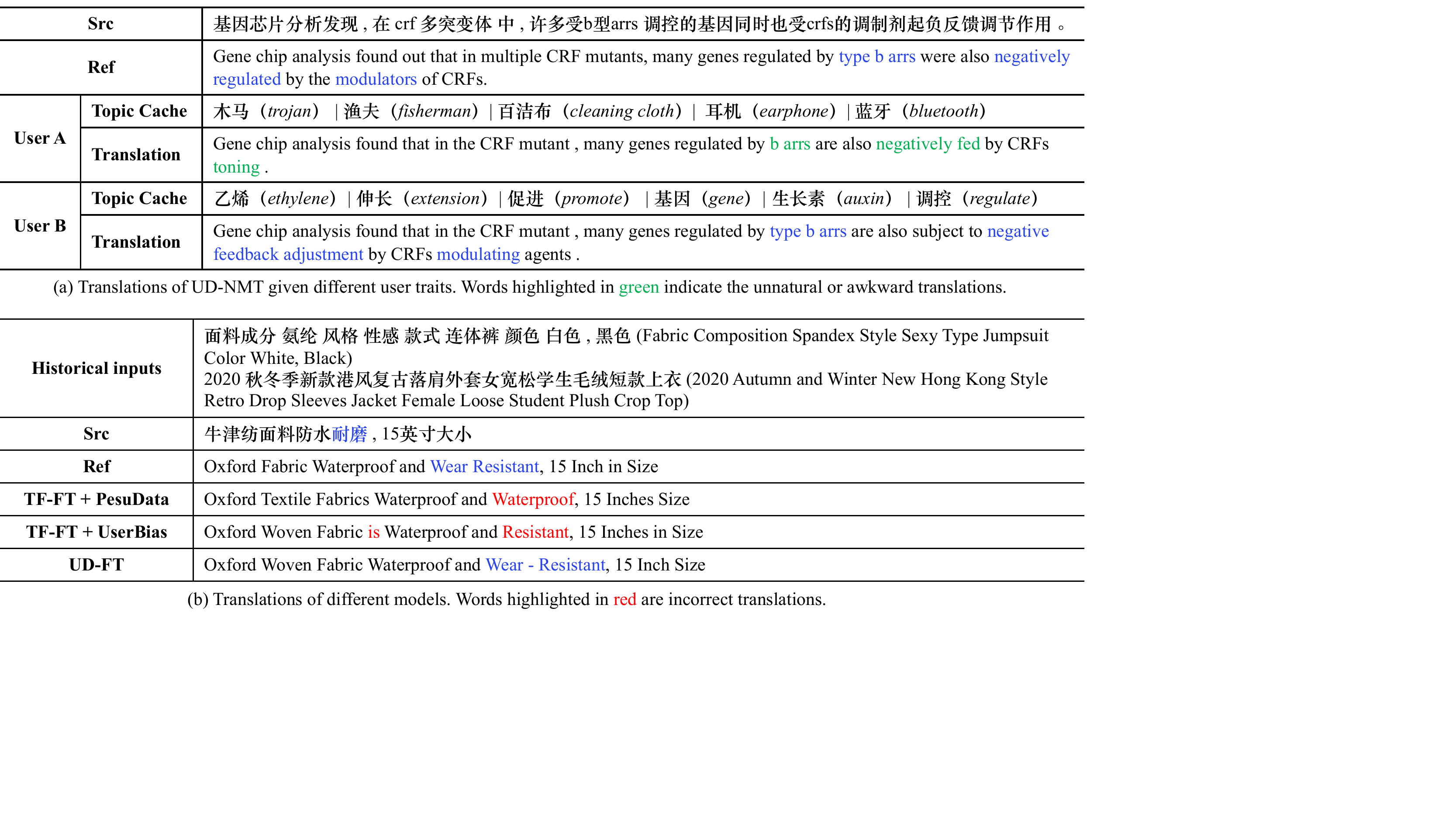}
\caption{
\label{fig_case}
Two examples of user-driven machine translation.
}
\end{figure*}


\subsection{Analysis of Contrastive Margin}
Inspired by \citet{acl/YangCLS19}, we argue that the contrastive learning may increase the prediction diversity of our model between users compared with using the MLE loss. To confirm this, we randomly sample 300 examples from the training dataset, and compute the following margin:
\begin{align}
\Delta = 
 \left[d^{(u^+)}\left(\cdot\right)
    \!-\! d^{(u^-)}\left(\cdot\right)\right]
    \!-\! \left[d^{(u^+)}_\mathbf{mle}\left(\cdot\right)
    \!-\! d^{(u^-)}_\mathbf{mle}\left(\cdot\right)\right],\notag
\end{align}
where $d^{(u^+)}(\cdot)$ is defined in Equation \ref{eqt_distance}. The definition of $d^{(u^+)}_\mathbf{mle}(\cdot)$ is the same with $d(\cdot)$, the only difference lies in that the NMT model is only trained by the conventional MLE loss. 
We find that $d(\cdot)$ has a larger margin than $ d_\mathbf{mle}(\cdot)$ on $88\%$ of sampled sentence pairs, with an average margin of 0.19. The results indicate again that the contrastive learning increases the translation diversity.

\subsection{Qualitative Analysis}
In order to intuitively understand how our cache module exactly affects the translations, we feed our model with the same current source sentence but different users, and display the 1-best translations generated by our model. 
As shown in the Figure \ref{fig_case} (a), 
our model is able to produce correct but diverse translations according to different topic caches. Moreover, it is interesting to observe that specific topic keywords such as \emph{``type b arr''}, \emph{``negatively regulated''} and \emph{``modulators''} are translated to synonymous but ``out-of-domain'' phrases if the topic cache does not conform to input sentence. On the contrary, the model conversely generates ``in-domain'' translation if the topic cache comes from the same topic of input sentence.

Besides, to further reveal the effect of user behavior, we provide an example in Figure \ref{fig_case} (b), which lists different translations by compared models for the same inputs. 
The historical inputs indicate that this user may be an 
apparel seller, since his historical inputs contain the product titles and descriptions of clothing. Thus, the keywords ``\emph{Wear Resistant}'' in the source sentence are correlated with this user. 
However, two baselines translate it to \emph{``Waterproof''} and \emph{``Resistant''}, respectively. Moreover, \emph{TF-FT + UserBias} generates a subject–verb–object structured sentence by adding the auxiliary verb ``is'', which does not conform to the expression habit of the product title.
By contrast, with the hint of the keywords in historical inputs, our \emph{UD-NMT} is able to produce suitable translation consistent with the topic preference of this user.

\begin{table}
\centering
\scalebox{0.93}{
\begin{tabular}{c|c}
\hline 
\textbf{Correlation Order} &\textbf{Proportion} \\
\hline
UD-NMT $>$ TF-FT + PesuData & 86\% \\
UD-NMT $>$ TF-FT + UserBias & 74\% \\
\hline
\end{tabular}}
\caption{The proportion of translations more related to historical inputs assessed by human translators. A $>$ B indicates the translations generated by A system is more correlated to history inputs. }
\label{tab_manual}
\end{table}

\subsection{Manual Evaluation}
To further find out weather the improvements of our model are contributed by user traits, we randomly sample 100 examples from the test dataset and ask the linguist experts to sort different systems according to the relevance between the generated translations and the historical input. The results in Table \ref{tab_manual} show that our model can generate translations more in line with history inputs than baseline models in most cases, proving that our method can make better use of user traits.

\section{Conclusion}

We propose user-driven NMT task, which aims to leverage user behavior to generate personalized translations. With the help of cache module and contrastive estimation, we successfully build an end-to-end NMT model that is able to capture potential user traits from their historical inputs and generate diverse translations under a zero-shot learning fashion. Furthermore, we contribute UDT-Corpus, which is the first Chinese-English parallel corpus annotated with user behavior. We expect our study can attract more attention towards this topic. 
It is a promising direction to explore other behavior in future, such as clickthrough and editing operations. 
Moreover, following recent advancements in domain adaptation for NMT, we plan to further improve our model via adversial training based knowledge transfer~\citep{DBLP:conf/emnlp/ZengSWLXYZ18,coling/YaoYZCL20,DBLP:journals/pami/SuZXWYL21} and dual knowledge transfer~\citep{DBLP:conf/emnlp/ZengLSGLYL19}.




\section*{Acknowledgments}
The project was supported by National Key Research and Development Program of China (No. 2020AAA0108004 and No. 2018YFB1403202), National Natural Science Foundation of China (No. 61672440), Natural Science Foundation of Fujian Province of China (No. 2020J06001), Youth Innovation Fund of Xiamen (No. 3502Z20206059), and the Fundamental Research Funds for the Central Universities (No. ZK20720200077). We also thank the reviewers for their insightful comments.

\bibliographystyle{acl_natbib}
\bibliography{acl2021}


\end{document}